# Flexibly-bounded Rationality and Marginalization of Irrationality Theories for Decision Making

Tshilidzi Marwala
South Africa
tmarwala@gmail.com

In this paper the theory of flexibly-bounded rationality which is an extension to the theory of bounded rationality is revisited. Rational decision making involves using information which is almost always imperfect and incomplete together with some intelligent machine which if it is a human being is inconsistent to make decisions. In bounded rationality, this decision is made irrespective of the fact that the information to be used is incomplete and imperfect and that the human brain is inconsistent and thus this decision that is to be made is taken within the bounds of these limitations. In the theory of flexibly-bounded rationality, advanced information analysis is used, the correlation machine is applied to complete missing information and artificial intelligence is used to make more consistent decisions. Therefore flexibly-bounded rationality expands the bounds within which rationality is exercised. Because human decision making is essentially irrational, this paper proposes the theory of marginalization of irrationality in decision making to deal with the problem of satisficing in the presence of irrationality.

## 1. Introduction

This paper is an extension of the paper on the theory of flexibly-bounded rationality which was proposed by Marwala (2013). Making decisions has been a complex task since time immemorial. In primitive society making decisions was muddled with superstitions and therefore quite irrational (Vyse, 2000; Foster and Kokko, 2009). Such irrationality in making decisions has even been speculated to occur in animals such as pigeons (Skinner, 1948). In essence, superstition which is irrational can be defined as supernatural causality where something is caused by another without them linked to one another. The idea of one event causing another without any connection between them whatsoever will be characterized as irrational thinking because it defies logic.

This paper proposes that the three fundamentals of decision making are information, correlation machine which is used to estimate missing data and causality, which relates inputs to outputs. As societies make attempt to move away from superstition and irrationality the concept of rational decision making comes to the fore. The concept of rationality has attracted many philosophers from different fields. Max Weber studied the role of rationality in social action (Weber, 1922) whereas Grayling studied the role of other social forces such as emotions on the quality or degree of rationality. Mosterin (2008) defined reason as a psychological facility and rationality as the optimizing strategy. In artificial intelligence a rational agent is that which aims to maximize its utility in making its decision and this is the optimization view of rationality.

Fundamentally, rational decision making is a process of reaching decisions through logic, information and in an optimized fashion (Nozick, 1993; Spohn, 2002). Decision making can be defined as a process through which a decision is reached. It usually involves many possible decisions outcomes and is said to be rational and optimal if it maximizes the utility that is derived from its consequences. The philosophical concept that states that the best course of action is that which maximizes utility is known as utilitarianism and was advocated by philosophers such as Jeremy Bentham and John Stuart Mills (Adams, 1976; Anscombe, 1958; Bentham, 2009; Mill, 2011).

There is a theory that has been proposed to achieve this desired outcome of a decision making of maximizing utility and this is the theory of rational choice and it states that one chooses a decision based on the product of the impact of the decision and its probability of occurrence (Allingham, 2002; Bicchieri, 2003). However, this theory was found to be inadequate because it does not take into account where the person will be relative to where they initially were before making a decision. Kahneman and Tversky (1979) extended the theory of rational choice by introducing the prospect theory which includes the reference position on evaluating the optimal decision (Kahneman, 2011).

Marwala (2013) proposed a generalized decision making process with two sets of input information which form the basis for decision making and options that are available for making such decisions (model) and the output being the decision being made. He further proposed that within this generalized decision making framework lie the statistical analysis device, correlation machine and a causal machine. Thus Mawala's model for decision making can be summarized as follows:
1. Optimization Device: e.g. maximizing utility and minimizing loss.
2. Decision making Device: In this paper we propose that this element contains statistical analysis device, correlation machine and causal machine.

The next section will describe in detail the concept of rational decision making, and the following section will describe the concept of bounded rationality which will then be followed by the concept of flexibly-bounded rationality. Thereafter, a generalized model for decision making is proposed followed by the theory of marginalization of irrationality for decision making process.

## 2. Rational Decision Making: A causal approach

Rational decision making in this paper is defined as a process through which a decision is made using some gathered information and some intelligence to make an optimized decision. It is deemed to be rational because it is based on evidence in the form of information. Making decisions in the absence of any information is deemed irrational. Rational decision making as illustrated in Figure 1 entails using data and a model (e.g. human brain) to make a decision. The relationship between data and the impact of the decision is used to improve or reinforce the learning process.

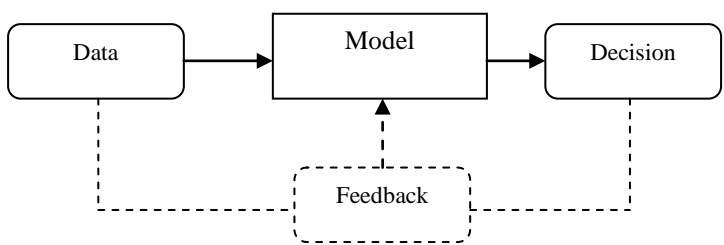

Figure 1 An illustration of a rational decision making system

Baker (2013) conducted a case study on rational decision making in a hybrid organization for a Southwestern Drug Court treatment program. Serrano González *et. al.* (2013) successfully applied rational decision making to identify optimal structure of transmissions systems for offshore wind farms. Salem *et. al.* (2013) applied successfully rational multi-criteria decision making to choose an operational plan for bridge rehabilitation. Multi-criteria optimization is an optimization of problem with more than one objective (Steuer, 1986; Köksalan *et. al.*, 1995; Zionts and Wallenius, 1976). As an example, one is given an option to buy a certain number of Microsoft stocks and General Electric stocks and the decision is how one combines these two stocks.

Chai *et. al.* (2013) reviewed how rational decision making has been applied successfully to choose a supplier selection whereas Badri *et. al.* (2013) applied rational decision making successfully to identify fireproofing requirements against jet fires. Other studies on the subject include studying a relationship between subjective memory and rational decision making (Hembacher and Ghetti, 2013) as well as successful application of rational decision making to economics (Gu *et. al.*, 2013).

## 3. Bounded-Rational Decision Making

One definition of rationality includes the concept of making an optimized decision. Practically this is not possible because of the limitation of the availability of information required to make such a decision and the limitation of the device to make sense of such incomplete decision (Simon, 1990). This is what is known as the theory of bounded rationality and was proposed by Nobel Laureate Herbert Simon (Simon, 1957; Tisdell, 1996). The implication of this observation on many economic systems is quite substantial (Simon, 1991). It questions the truthfulness of the theory of the efficient market hypothesis especially given the fact that participants in the market are at best only rational within the bounds of limited information and limited model to make sense of the information.

Lee (2013) built a model for analyzing an economy using bounded rational agents. It was observed that rational agents amplify the price bubble cooperation amongst rational agents. Gama (2013) successfully applied the concept of bounded rationality to investigate the behavior of stream mining procedures and proposed ubiquitous stream mining and self-adaption models. Jiang *et. al.* (2013) successfully applied bounded rationality in evolution game analysis of water saving and pollution prevention. Jin *et. al.* (2013) successfully applied the theory of bounded rationality and game theory to construct a computer virus propagation model and observed that the proposed model was able to forecast the propagation of computer virus.

Yao and Li (2013) surmised that the concept of bounded rationality can be viewed as a basis of loss aversion and optimism and in this regard studied psychological adaptation within the context of incomplete information. They observed that loss aversion and optimism arise when the degree of data incompleteness exceeds a particular threshold. In addition, they observed that loss aversion and optimism develop to be more noticeable when information is sparser. They then concluded that psychological biases benefit from apparent information incompleteness when value creation is considered.

Stanciu-Viziteu (2012) simulated the behavior of sharks to model financial market where investors are embodied by hungry sharks. The results obtained indicated that sharks resembling investors coordinate and produce equilibrium under rational expectations.

Aviad and Roy (2012) applied the concept of bounded rationality to build a decision support system and applied this to identify feature saliency in clustering problems. Bounded rationality was applied to deal with the problem of the availability of target attribute by using an S-shaped function as a saliency measure to characterize the end user's logic to identify attributes that describe each prospective group.

Murata *et. al.* (2012) applied the concept of bounded rationality to analyze the relationship between group heuristics and cooperative behavior. They observed that group consciousness can assist to stimulate actively mutual cooperation.

The literature review studied above clearly indicates that the theory of bounded rationality is a powerful tool of analysis that has found usage in many diverse areas. It should be noted that the theory of bounded rationality has not replaced the theory of rationality which was described in earlier sections. What this theory has done is to put limitations or constraints on the applicability of the theory of rationality. The theory of rationality which was described in Figure 1 can thus be updated to construct the theory of bounded rationality as described in Figure 2

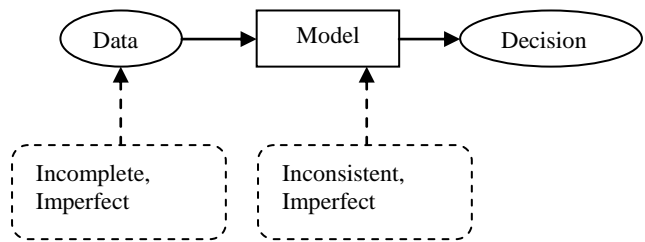

Figure 2 Illustration of the theory of bounded rationality

Figure 2 illustrates that on making decisions rationally, certain considerations must be taken into account and these are that the data that is used to make such decisions are never complete and are subject to measurement errors and are thus imperfect and incomplete. The second aspect that should be observed is that the model which takes data and translates these into decisions is also imperfect. If for example, this model is a human brain then it is inconsistent in the sense that it may be influenced by other factors such as whether the individual concerned is hungry, or angry. Herbert Simon coined a term *satisficing,* thereby hybridizing the terms satisfying and sufficing, which is the concept of making an optimized decision under the limitations that the data used in making such decisions are imperfect and incomplete while the model used to make such decisions is inconsistent and imperfect.

Now that we have described the theory of bounded rationality, which resulted in the limitations of the applicability of the theory of rational decision making, the next section describes the theory of flexibly-bounded rationality which is theory proposed in this paper.

## 4. Flexibly-bounded Rational Decision Making

This section proposes the theory of flexibly-bounded rationality which is shown in Figure 3. In order to understand the theory of flexibly-bounded rationality, it is important to state few propositions and these are:
 1. Rational decision making is a process of making optimized decisions based on logic and scientific thinking based on information. The issue of optimized decision becomes even necessary given the migration of decision making towards using intelligent machines.
 2. The process of rationality is indivisible. In other words, you cannot be half rational and half irrational. If you are half rational and half irrational you are being irrational. The concept of the indivisibility of rationality and the impact of this on many facets of life is a powerful concept that requires further investigation.
3. The principle of bounded rationality does not truncate the theory of rationality but merely specifies the bounds within which the principle of rationality is applied.

With advances in information processing techniques, enhanced theories of autoassociative machines and advances in artificial intelligence methods, it has become necessary to revise the theory of bounded rationality. The fact that information which is used to make decisions is imperfect because of factors such as measurement errors can be partially corrected by using advanced information analysis methods and the fact that some of the data that are missing and, thereby, incomplete can be partially completed using missing data estimation methods and the fact that a human brain which is influenced by other social and physiological factors can be substituted for by recently developed artificial intelligence machines implies that the bounds under which rationality is exercised can be shifted and thus bounded rationality can now be transformed into the theory of flexibly-bounded rationality.

Tsang (2008) proposed that computational intelligence determines effective rationality. What Tsang implies is that there is a degree of rationality, a situation which is not true. Rationality cannot be quantified, it is either there or absent. The model of flexibly-bounded rationality can thus be expressed as shown in Figure 3.

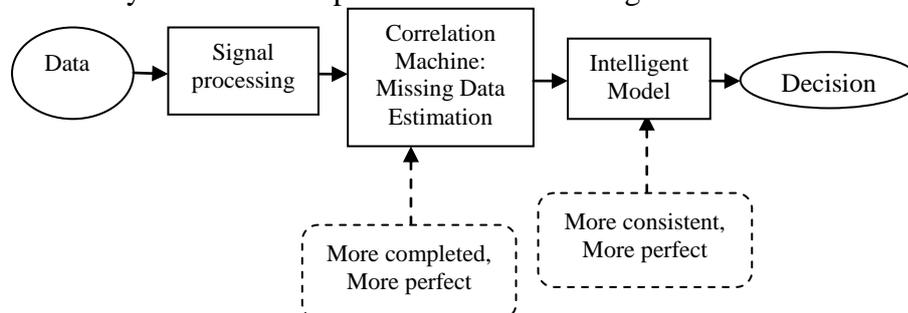

Figure 3 Illustration of the theory of flexibly-bounded rationality.

The implications of the model in Figure 3 to many disciples such as economics, political science and social science are extensive. It goes to the heart of the relationship between rationality and technology. Basically, modern technology now allows us to update the limiting concept of bounded rationality to a less limited concept of flexibly-bounded rationality. The implication of the concept of the notion of indivisibility of rationality and the application of this profound notion to many areas such as economics, sociology and political science is far reaching and requires further investigation. The schematic illustration of flexibly-bounded rationality as compared to bounded rationality is shown in Figure 4.

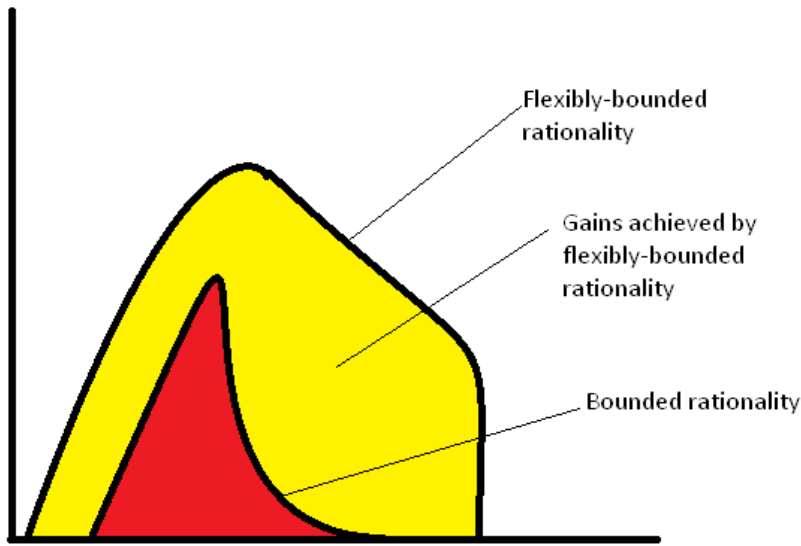

Figure 4 Comparison between flexibly-bounded rationality and bounded rationality

Figure 4 demonstrates that the theory of flexibly-bounded rationality only expands the bounds within which the principle of rationality is applied. It also shows that the theory of bounded rationality only prescribes the bounds within which the principle of rationality is applied.

The next section investigates whether rationality is divisible while the subsequent subsections describe advanced information processing, missing data estimation methods and artificial intelligence techniques which are elements that are proposed in this paper to take the theory of bounded rationality into a theory of flexibly-bounded rationality.

## 5. Is Rational Decision Making Process Necessarily Optimal?

Our definition of rationality includes three key concepts and these are: logic (based on scientific principles), information (it should be based on information) and optimization (it should be optimized). The notion that for a decision to be rational it has to be optimized has a far reaching implication on human decision making. It essentially implies that on the whole human beings are fundamentally irrational beings. An optimized decision essentially implies that such a decision results in the maximization of utility (DeGroot, 1970; Berger, 1980). If for example, if one needed to go from Johannesburg to Vancouver then she decides to go to London then to Nairobi and then to Vancouver for no other reason except to try to reach Vancouver from Johannesburg then this behaviour will be considered irrational. The reason why this will be considered irrational is because it is not an optimized decision and, therefore, wastes sources. Wasting resources is not rational. As we move from human decision making into decision making by artificial intelligent machines, it is vital that we capacitate these machines with the ability to make logical and optimized decisions.

## 6. Can rationality be divided?

The fact that the theory of bounded rationality as well as the theory of flexibly-bounded rationality merely prescribes the bounds within which rationality is exercised implicitly implies that the principle of rationality is not divisible. The question that ought to be asked, therefore, is whether rationality can be divided. In other words can a person act half rationally? To illustrate this point, suppose a young man named Masindi who was just diagnosed with a heart problem decides to consult an astrologer who throws down some bones and prescribes that Masindi needs to climb Mountain Luvhola in order to be healed. If Masindi accepts this advice we will declare him to be thinking irrationally. Suppose Masindi then decides to go and consult a cardiologist to deal with his heart problem, then we will declare him to be rational. If Masindi decides to go and see both the astrologer and a cardiologist, then is he acting rationally, irrationally or half rationally? Clearly, Masindi is acting irrationally because his process of dealing with his process is illogical and not optimized. This paper surmises that if a person is making a decision even a small dose of irrationality renders the whole decision making process irrational. Rationality is surmised to be indivisible in this paper. Masindi either acts rationally or irrationally and not in between the two.

## 7. Advanced Information Processing

This section describes advanced information processing techniques that are required in order to use data to make decisions within the context of flexibly-bounded rationality possible. These techniques are the time domain, frequency domain and time-frequency domain frameworks.

*A. Time Domain Data*
When decisions are made, data can be used in the time domain to make decisions. In simple terms, data in the time domain means that the data set is used exactly as it was sampled at a given time. For example, if one wants to use the data of temperature per unit time today to predict the average temperature tomorrow one can observe a series of data today to make tomorrow's predictions.

Coillet *et. al.* (2013) successfully applied the time domain methods to analyze optoelectronic oscillators based on whispering-gallery mode resonators. Szurley *et. al.* (2013) successfully applied the time domain technique to enhance binaural speech. Pereira *et. al.* (2013) applied the time domain data from nuclear magnetic resonance and chemometrics to classify fresh plums according to sweetness. Herckenrath *et. al.* (2013) successfully applied electromagnetic data in the time domain methods to calibrate salt water intrusion model. Paulraj *et. al.* (2013) successfully applied the time domain data to classify moving vehicle.

*B. Frequency Doman*
The basis of frequency domain method is the Fourier series and it fundamentally states that every periodic function can be approximated by a Fourier series which is a combination of series of cosines and sine functions of various frequencies (Fourier, 1822; Boashash, 2003; Bochner and Chandrasekharan, 1949; Bracewell, 2000). This means that a signal can be characterized by a series of cycles with different amplitudes and frequencies.

Qu *et. al.* (2013) successfully applied frequency domain methods to filter gradient image for contour detection whereas Oh *et. al.* (2013) applied hydrodynamic load in the frequency domain to analyze offshore wind turbine. Mohapatra *et. al.* (2013) successfully applied electromechanical disturbances in the frequency domain to study electric power systems whereas Zhang *et. al.* (2013) successfully applied optical coherence tomography in the frequency domain to detect glaucoma.

*C. Time-Frequency Domain*
Time-frequency methods are approaches where it is possible to view data in both the time and frequency domains simultaneously. There are different types of time-frequency methods and these comprise Short-time Fourier transform (Jont, 1977; Jacobsen and Lyons, 2003), wavelet transform (Chui, 1992: Akansu and Haddad, 1992), Wigner distribution function (Wigner, 1932; Qian and Chen, 1992) and Gabor distribution (Daugman, 1980; Jones and Palmer, 1986).

Ma *et. al.* (2013) successfully applied time-frequency analysis for blind source separation whereas Le and Paultre (2013) successfully applied time-frequency analysis for modal identification. Zhang *et. al.* (2013) successfully applied time-frequency analysis for real world source separation whereas Jiang *et. al.* (2013) successfully applied time-frequency analysis for fault diagnosis. Finally, Heideklang and Ivanova (2013) successfully applied time-frequency analysis to extract features from brain signals.

## 8. Missing Data Estimation

One aspect of the theory of flexibly-bounded rationality is that it uses missing data estimation techniques to reduce information that is not observed or known. In this section we describe missing data estimation method as discussed by Marwala (2009). Missing data estimation method is technique that is used to estimate data and thus information that is missing. It necessitates mathematical models that describe interrelationships that exist amongst the variables. There are many methods that have been proposed to approximate missing data. One of these is one which successfully applied autoassociative neural networks to capture variables' interrelationships and genetic algorithms to identify missing values (Abdella and Marwala, 2006).

The other technique that has been used is the one that uses a hybrid of the principal component analysis and neural networks to capture the interrelationships that exist amongst the data and genetic algorithm to estimate missing values (Mistry *et. al.,* 2009). Nelwamondo *et. al.* (2007) successfully applied two missing data estimation techniques which were the Gaussian Mixture Models trained using the Expectation Maximization (EM) algorithm and autoassociative neural networks which used particle swarm optimization to estimate missing values. Nelwamondo and Marwala (2005) proposed the use of rough set theory to estimate missing values. Nelwamondo *et. al.* (2013) successfully applied dynamic programming and autoassociative neural network method to estimate missing data whereas Shi *et. al.* (2013) successfully applied Bayesian principal component analysis and iterative local least squares to estimate missing values in microarray data. Pati and Das (2012) successfully applied similarity measurement for missing value estimation in microarray data whereas Luy and Saray (2012) successfully applied back-propagation algorithms to estimate missing wind speed data. Garnier *et. al.* (2012) successfully estimated missing data for energy resources management in tertiary buildings whereas Bose *et. al.* successfully applied a unique interpolation based missing value estimation method to predict missing values in microarray gene expression data.

This paper estimates missing data using autoassociative multi-layer perceptron neural networks to capture the model that describes the data and genetic algorithms to estimate the missing values. This technique is described in detail by Marwala (2009).

## 9. Intelligent Machines

The core of the theory of flexibly-bounded rationality is that it assumes that an artificial intelligent machine is used for decision making process because it is deemed to be better than the human brain. There are many different types of artificial intelligence methods that can be used to make decisions and one of these is neural network which is used in this paper. Neural network is an artificial intelligence technique which is inspired by how a human brain in structured which is used to learn patterns from data (Patel and Marwala, 2006). There are many types of neural networks types and these include the radial basis function and the multi-layer perceptron neural network which is used in this paper. Ensemble of neural networks and also called a committee of networks have been used to relate two sets of data (Marwala, 2001; Marwala *et. al.,* 2011). Other types of artificial intelligence techniques that can be used include the Gaussian mixture models (Marwala *et. al.,* 2006; Bishop, 2006), hidden Markov model (Baum and Petrie, 1966), fuzzy logic (Zadeh, 1996), fuzzy ARTMAP (Carpenter *et. al.,* 2003; Vilakazi and Marwala, 2006), rough sets (Tettey *et. al.,* 2007), and support vector machine (Vapnik, 1995; Schölkopf and Smola, 2002).

## 10. What is Irrationality?

In this paper we define rational decision making as a process of making decisions based on the process that is logical and is premised on information and is optimized. Given this definition, then what is irrational decision making? A decision making process becomes irrational if it is not based on information (evidence) or it is illogical or it is un-optimized. In the earlier part of this paper we did conclude that human beings are essentially irrational beings.

## 11. Irrational Decision Making and the Theory of Marginalization of Irrationality in Decision Making

The sobering conclusion that human beings are most often than not irrational poses an additional dilemma on how they do manage to satisfice even when they are acting irrationally. I will go back to the patient Masindi who after being diagnosed with a serious heart problem he consulted an astrologer. On the morning of day 1 he goes to the astrologer to seek help and thereafter goes straight to see a cardiologist who examines him and tells him that he is going for an operation the following day. The following day before going to the operation room he consults an astrologer to strengthen the chances of success of the operation. Now, in the process of dealing with his heart problem, is Masindi being rational? The three aspects of rationality discussed in this paper are logic, information and optimization. Masindi is not being logical and certainly his decision is not optimized and, therefore, is irrational. Now under these circumstances, how does one make decisions that are not rational but are still satisficing? In this regard, we propose the theory of marginalization of irrational power to make satisficed decisions. The premise of this theory is that if there is so much irrationality in human decision making then why do we have societies that seem to function?

If we consider the Masindi example, then we can see that in the process of dealing with the heart problem, he pursued two approaches one which we deem rational (going to the cardiologist) and another one that we deem irrational (going to an astrologer) but collectively the whole process of dealing with the problem is irrational because it is neither optimal nor logical. In this case a satisficed decision will only emerge if the irrational aspect of the process that Masindi follows is marginalized in favour of the rational decision of seeing a cardiologist. It may quite be possible that the astrologer could have recommended the ingestion of lithium battery which could have been fatal and therefore the irrational aspect of the process could have overwhelmed the rational aspect.

Given this example, how then do we generalize this theory of the marginalization of irrationality for decision making? The steps that ought to be followed to make a decision when the process is not rational are as follows:

1. Collect the information needed to make a decision.
2. Outline a process to be followed to make such as a decision.
3. Divide each step of the process into whether the step is rational or irrational.
4. Assess whether the irrational aspects of the process are marginalizable when compared to the rational steps.
5. If the irrational aspects of the steps are marginalizable then the solution is satisficing whereas if these aspects are not marginalizable then the solution is not satisficing.

Clearly satisficing an irrational decision requires the assessment as to whether the irrational aspect of an irrational process of decision making is marginalizable. How do we achieve this? In this paper we borrow from the field of information theory the concept of signal to noise ratio and adapt it to the concept of the ratio of the power of rationality to the power of irrationality (Raol, 2009). One difficult way is to identify the causal loop between the irrational part of the process and a decision and evaluate the impact of this on the decision and compare this to the impact of the rational part of the decision process on the decision to be able to calculate the power of rationality to the power of irrationality ratio.

To illustrate this point, I will invoke the story of Nongqawuse the Xhosa prophet who advised her community that if they kill their cattle and crops then God will drive the British into the sea (Peires, 1989; Stapleton, 1991; Welsh, 2000). Indeed the overwhelming majority which is estimated to be approximately 95% of community members killed the cattle and crops. From this, the power of rationality to the power of irrationality is approximately 0.05 and therefore there was no chance that this decision process could have resulted in a satisficed outcome. The process this community followed had both rational and irrational aspects but was in the whole irrational. The result of this was famine and deaths leading to a population decline from 127,000 to 27,000. Using the theory of marginalization of irrationality, the outcome was not satisficing and the reason for this is because the irrational aspects of this irrational process could not be marginalized.

The theory of marginalization of information is also applicable to any decision making process where the information that is needed to make such a decision is not complete and there is no mechanism of completing either partially or optimistically fully the missing information. In this situation, the decision maker simply ignores or marginalizes the missing information and continues to make a decision. The efficacy of this decision will depend on the ratio of the power of observed information to the power of the missing information just exactly as is the case in information theory where the usefulness of the information depends on the signal to noise ratio.

## 13 Conclusions

In this paper the concept of flexibly-bounded rationality is introduced as an extension to the notion of bounded rationality. It is based on the principle of rationality but is realized by using advanced signal processing, missing data estimation methods and artificial intelligence techniques. Flexibly-bounded rationality expands the bounds within which rational decision making process is exercised and, thereby, increases the probability of making accurate decisions when compared to the theory of bounded rationality. In addition, the theory of the marginalization of irrationality to satisfice a decision was also proposed. This theory was applied to analyze a historical account of an event in South Africa and was able to show why the outcome was not satisficed.

## References

Abdella M. and Marwala T. "The use of genetic algorithms and neural networks to approximate missing data in database", Computing and Informatics, Vol.24, pp. 1001-1013, 2006

Adams, R.M. "Motive Utilitarianism", The Journal of Philosophy,Vol.73,No.14, 1976

Akansu, A.N.; Haddad, R.A. "Multiresolution Signal Decomposition: Transforms, Subbands, Wavelets", San Diego: Academic Press. ISBN 978-0-12-047141-6, 1992

Allingham, M. "Choice Theory: A Very Short Introduction", Oxford, 2002

Anscombe, G. E. M. "Modern Moral Philosophy", Philosophy, Vol.33,No.124, January 1958

Arab-Chamjangali, M. "Modelling of cytotoxicity data (CC50) of anti-HIV 1-[5-chlorophenyl) sulfonyl]-1H-pyrrole derivatives using calculated molecular descriptors and Levenberg-Marquardt artificial neural network", Chemical Biology and Drug Design, Vol.73,No.4, pp. 456-465, 2009

Atencia, M., Joya, G., Sandoval, F. "Modelling the HIV-AIDS Cuban epidemics with Hopfield neural networks", Lecture Notes in Computer Science (including subseries Lecture Notes in Artificial Intelligence and Lecture Notes in Bioinformatics), Vol.2687, pp. 449-456, 2003

Aviad, B., Roy, G. "A decision support method, based on bounded rationality concepts, to reveal feature saliency in clustering problems", Decision Support Systems, Vol.54, No.1, pp. 292-303, 2012

Badri, N., Rad, A., Kareshki, H., Abdolhamidzadeh, B., Parvizsedghy, R., Rashtchian, D. "A risk-based decision making approach to determine fireproofing requirements against jet fires", Journal of Loss Prevention in the Process Industries, Vol.26, No.4, pp. 771-781, 2013

Baker, K.M. "Decision Making in a Hybrid Organization: A Case Study of a Southwestern Drug Court Treatment Program", Law and Social Inquiry, Vol.38, No.1, pp. 27-54, 2013

Baum, L. E.; Petrie, T. "Statistical Inference for Probabilistic Functions of Finite State Markov Chains", The Annals of Mathematical Statistics, Vol.37, No.6, pp.1554–1563, 1966, 20

Bentham, J. "An Introduction to the Principles of Morals and Legislation (Dover Philosophical Classics)", Dover Publications Inc, 2009

Berger, J.O. "Statistical Decision Theory and Bayesian Analysis", Springer-Verlag, Heidelberg, 1980.Bicchieri, C. "Rationality and Game Theory", in The Handbook of Rationality, The Oxford Reference Library of Philosophy, Oxford University Press, 2003

Bishop, C. "Pattern recognition and machine learning", New York: Springer, 2006

Boashash, B., ed. "Time-Frequency Signal Analysis and Processing: A Comprehensive Reference", Oxford: Elsevier Science, 2003

Bochner S., Chandrasekharan K, "Fourier Transforms", Princeton University Press, 1949

Bochner S., Chandrasekharan K. "Fourier Transforms", Princeton University Press, 1949

Bose, S., Das, C., Dutta, S., Chattopadhyay, S. "A novel interpolation based missing value estimation method to predict missing values in microarray gene expression data", Proceedings of the 2012 International Conference on Communications, Devices and Intelligent Systems, CODIS 2012, art. no. 6422202, pp. 318-321, 2012

Bracewell, R. N. "The Fourier Transform and Its Applications (3rd ed.) ", Boston: McGraw-Hill, 2000

Carpenter, G.A. & Grossberg, S. "Adaptive Resonance Theory", In Michael A. Arbib (Ed.), The Handbook of Brain Theory and Neural Networks, Second Edition (pp. 87-90). Cambridge, MA: MIT Press, 2003

Chai, J., Liu, J.N.K., Ngai, E.W.T. "Application of decision-making techniques in supplier selection: A systematic review of literature", Expert Systems with Applications, Vol.40, No.10, pp. 3872-3885, 2013

Chui, C.K. "An Introduction to Wavelets. San Diego", Academic Press,1992

Coillet, A., Henriet, R., Salzenstein, P., Huy, K.P., Larger, L., Chembo, Y.K. "Time-domain dynamics and stability analysis of optoelectronic oscillators based on whispering-gallery mode resonators", IEEE Journal on Selected Topics in Quantum Electronics art. no. 6477074, Vol.19,No.5, 2013

Daugman, J.G. "Two-dimensional spectral analysis of cortical receptive field profiles", Vision Res., PMID 7467139, Vol.20,No.10,pp.847–56, 1980

DeGroot, M. "Optimal Statistical Decision", McGraw-Hill, New York, 1970.

Dybała, J., Radkowski, S. "Reduction of Doppler effect for the needs of wayside condition monitoring system of railway vehicles", Mechanical Systems and Signal Processing, Vol.38, No.1, pp. 125-136, 2013

Fernández, M., Caballero, J. "Modeling of activity of cyclic urea HIV-1 protease inhibitors using regularized-artificial neural networks", Bioorganic and Medicinal Chemistry, Vol.14, No.1, pp. 280-294, 2006

Foster; K.R. and Kokko H. "The evolution of superstitious and superstition-like behaviour", Proceedings of the Royal Society B: Biological Sciences, Vol.276,No.1654,pp.31–7, 2009

Fourier, J. B. "Théorie Analytique de la Chaleur, Paris: Chez Firmin Didot", père et fils, . 1822

Gama, J. "Data stream mining: The bounded rationality", Informatica (Slovenia),Vol.37,No.1, pp. 21-25, 2013

Garcia, R.F., Rolle, J.L.C., Gomez, M.R., Catoira, A.D. "Expert condition monitoring on hydrostatic self-levitating bearings", Expert Systems with Applications, Vol.40,No.8, pp. 2975-2984, 2013

Garnier, A., Eynard, J., Caussanel, M., Grieu, S. "Missing data estimation for energy resources management in tertiary buildings", 2nd International Conference on Communications Computing and Control Applications, CCCA 2012, art. no. 6417902, 2012

Gu, J., Bohns, V.K., Leonardelli, G.J. "Regulatory focus and interdependent economic decision-making", Journal of Experimental Social Psychology, Vol.49,No.4, pp. 692-698, 2013

Heideklang, R., Ivanova, G. "A novel flexible model for the extraction of features from brain signals in the time-frequency domain", International Journal of Biomedical Imaging, 2013, art. no. 759421, 2013

Hembacher, E., Ghetti, S. "How to bet on a memory: Developmental linkages between subjective recollection and decision making", Journal of Experimental Child Psychology, Vol.115, No.3, pp. 436-452, 2013

Herckenrath, D., Odlum, N., Nenna, V., Knight, R., Auken, E., Bauer-Gottwein, P. "Calibrating a Salt Water Intrusion Model with Time-Domain Electromagnetic Data", Ground Water, Vol.51,No.3, pp.385-397, 2013

Hung, S.-S., Chang, H.-C., Kao, P.-T., Lin, S.-L. "Automatic monitoring and control for green house environment based on optimal plant growing condition", Advanced Science Letters,Vol.19,No.10, pp. 2895-2900, 2013

Jacobsen E. and Lyons R, "The sliding DFT, Signal Processing Magazine", Vol.20, No.2, pp.74–80, 2003

Jiang, R., Xie, J., Wang, N., Li, J. "Evolution game analysis of water saving and pollution prevention for city user groups based on bounded rationality", Shuili Fadian Xuebao/Journal of Hydroelectric Engineering, Vol.32, No.1, pp. 31-36, 2013

Jiang, Z., Jiao, W., Meng, S. "Fault diagnosis method of time domain and time-frequency domain based on information fusion", Applied Mechanics and Materials, Vol.300-301, pp. 635-639, 2013

Jin, C., Jin, S.-W., Tan, H.-Y. "Computer virus propagation model based on bounded rationality evolutionary game theory", Security and Communication Networks, Vol.6,No.2, pp. 210-218, 2013

Jones J.P. and Palmer L.A. "An evaluation of the two-dimensional gabor filter model of simple receptive fields in cat striate cortex", J. Neurophysiol., Vol.58,No.6,pp.1233-1258, 1987

Jont B.A. (1977). "Short Time Spectral Analysis, Synthesis, and Modification by Discrete Fourier Transform", IEEE Transactions on Acoustics, Speech, and Signal Processing. ASSP-25, No.3, pp.235–238.

Kahneman, D., "Thinking, Fast and Slow" Macmillan, New York, 2011

Kahneman, D., Tversky, A. "Prospect Theory: An Analysis of Decision Under Risk", Econometrica, Vol.47, No.2, pp. 263–291, 1979

Köksalan, M.M. and Sagala, P.N.S., M. M.; Sagala, P. N. S. "Interactive Approaches for Discrete Alternative Multiple Criteria Decision Making with Monotone Utility Functions", Management Science,Vol.41,No.7,pp1158–1171, 1995

Le, T.-P., Paultre, P. "Modal identification based on the time-frequency domain decomposition of unknown-input dynamic tests", International Journal of Mechanical Sciences, Vol.71, pp. 41-50, 2013

Lee, I.H. "Speculation under bounded rationality", Journal of Economic Theory and Econometrics, Vol.24,No.1, pp. 37-53, 2013

Leke B.B, Marwala T., Tim T., Lagazio M. "Prediction of HIV Status from Demographic Data Using Neural Networks", Proceedings of the IEEE International Conference on Systems, Man and Cybernetics, Taiwan, pp. 2339-2344, 2006

Luy, M., Saray, U. "Wind speed estimation for missing wind data with three different backpropagation algorithms", Energy Education Science and Technology Part A: Energy Science and Research, Vol.30, No.1, pp. 45-54, 2012

Ma, X., Ding, S., Wei, L., Yang, J. "Blind identification of underdetermined mixing matrix and source separation by finding and solving a row echelon-like form of system in the time-frequency domain", ICIC Express Letters, Part B: Applications, Vol.4,No.3, pp. 739-746, 2013

Marivate, V. and Marwala, T. "Relational networks for HIV classification", Proceedings of the IASTED Africa Conference on Modelling and SimulationEditor: F.J. Ogwu, pp. 275-279, 2008

Marwala T. "Semi-bounded Rationality: A model for decision making", arXiv:1305.6037, 2013.

Marwala T and Crossingham B. "Neuro-rough models for modelling HIV", Proceedings of the IEEE International Conference on Man, Systems and Cybernetics, pp. 3089-3095, 2008

Marwala T. "Computational Intelligence for Missing Data Imputation, Estimation and Management: Knowledge Optimization Techniques", Information Science Reference Imprint, IGI Global Publications, New York, 2009.

Marwala T. "Condition Monitoring Using Computational Intelligence Methods", Springer-Verlag, Heidelberg, 2012.

Marwala T. "Finite element updating using wavelet data and genetic algorithm", American Institute of Aeronautics and Astronautics, Journal of Aircraft, Vol.39, pp.709-711, 2002

Marwala T. "Probabilistic fault identification using a committee of neural networks and vibration data", American Institute of Aeronautics and Astronautics, Journal of Aircraft, V0l.38, pp.138-146, 2001

Marwala T., de Wilde P., Correia L., Mariano P., Ribeiro R., Abramov V., Szirbik N., Goossenaerts J, "Scalability and optimisation of a committee of agents using genetic algorithm", In Proceedings of the International Symposia on Soft Computing and Intelligent Systems for Industry, Scotland, 2001.

Marwala T., Mahola U and Nelwamondo F. "Hidden Markov models and Gaussian mixture models for bearing fault detection using fractals", In the Proceedings of the IEEE International Joint Conference on Neural Networks, BC, Canada, ISBN: 0-7803-9489-5, pp. 5876-5881, 2006

Mill, J.S. "A System of Logic, Ratiocinative and Inductive (Classic Reprint)", Oxford University Press, 2011

Mistry J., Nelwamondo F.V., Marwala T. "Missing Data Estimation using Principal Component Analysis and Autoassociative Neural Networks", Journal of Systemics, Cybernetics and Informatics, Vol.7,No.3, pp. 72-79 , 2009

Mistry, J.; Nelwamondo, F.V.; Marwala, T. "Investigating a Predictive Certainty measure for Ensemble Based HIV Classification", IEEE International Conference on Systems, Computational Cybernetics, 2008. ICCC 2008, pp. 231-236, 27-29 Nov. 2008

Mohamed A.K, F. Nelwamondo V and Marwala T. "Estimation of missing data: Neural networks, principal component analysis and genetic algorithms", Proceedings of the 12th World Multi-Conference on  Systemics, Cybernetics and Informatics: WMSCI 2008, Orlando, Florida, USA, pp. 36-41, 29th June –2nd July

Mohapatra, S., Zhu, H., Overbye, T.J. "Frequency-domain analysis of electromechanical disturbances in electric power systems", 2013 IEEE Power and Energy Conference at Illinois, PECI 2013, art. no. 6506063, pp. 230-237, 2013

Murata, A., Kubo, S., Hata, N. "Study on promotion of cooperative behavior in social dilemma situation by introduction of bounded rationality - Effects of group heuristics on cooperative behavior", Proceedings of the SICE Annual Conference, art. no. 6318444, pp. 261-266, 2012

Nelwamondo F.V. and Marwala T. "Rough set theory for the treatment of incomplete data", Proceedings of the IEEE Conference on Fuzzy Systems, pp. 338-343, 2007

Nelwamondo F.V., Mohamed S. and Marwala T. "Missing Data: A Comparison of Neural Network and Expectation Maximisation Techniques", Current Science, Vol. 93, No.11, pp. 1514-1521, 2007

Nelwamondo, F.V., Golding, D., Marwala, T. "A dynamic programming approach to missing data estimation using neural networks" Information Sciences, Vol.237, pp. 49-58, 2013

Nozick, R. "The Nature of Rationality", Princeton: Princeton University Press, 1993

Oh, K.-Y., Kim, J.-Y., Lee, J.-S. "Preliminary evaluation of monopile foundation dimensions for an offshore wind turbine by analyzing hydrodynamic load in the frequency domain", Renewable Energy, Vol.54, pp. 211-218, 2013

Patel P and Marwala T.  " Neural networks, fuzzy inference systems and adaptive-neuro fuzzy inference systems for financial decision making", Lecture Notes in Computer Science, Springer-Verlag, Berlin Heidelberg, Vol.4234, pp. 430-439,  2006

Pati, S.K., Das, A.K. "Missing value estimation of microarray data using similarity measurement", Lecture Notes in Computer Science (including subseries Lecture Notes in Artificial Intelligence and Lecture Notes in Bioinformatics), 7677 LNCS, pp. 602-610, 2012

Paulraj, M.P., Adom, A.H., Sundararaj, S. "Classification of moving vehicle using multi-frame time domain features", 7th International Conference on Intelligent Systems and Control, ISCO 2013, art. no. 6481211, pp. 529-533, 2013

Peires, J.B., "The Dead Will Arise: Nongqawuse and the Great Xhosa Cattle-Killing Movement of 1856-7", Indiana University Press, 1989.

Pereira, F.M.V., Carvalho, A.D.S., Cabeça, L.F., Colnago, L.A. "Classification of intact fresh plums according to sweetness using time-domain nuclear magnetic resonance and chemometrics", Microchemical Journal, 108, pp. 14-17, 2013

Qian S. and Chen D. "Joint Time-Frequency Analysis: Methods and Applications", Chap. 5, Prentice Hall, N.J., 1996.

Qu, Z.-G., Wang, P., Gao, Y.-H., Wang, P., Shen, Z.-K. "Frequency domain filtering of gradient image for contour detection", Optik, Vol.124, No.13, pp. 1398-1401, 2013

Raol, J.R. "Multi-sensor data fusion: Theory and Practice" CRC Press, 2009.

Salem, O.M., Miller, R.A., Deshpande, A.S., Arurkar, T.P. "Multi-criteria decision-making system for selecting an effective plan for bridge rehabilitation", Structure and Infrastructure Engineering, Vol.9,No.8, pp. 806-816, 2013

Schölkopf, B. and Smola, A.J. "Learning with Kernels", MIT Press, Cambridge, MA, 2002

Serrano González, J., Burgos Payán, M., Riquelme Santos, J. "Optimum design of transmissions systems for offshore wind farms including decision making under risk", Renewable Energy, Vol.59, pp. 115-127, 2013

Shi, F., Zhang, D., Chen, J., Karimi, H.R. "Missing value estimation for microarray data by Bayesian principal component analysis and iterative local least squares", Mathematical Problems in Engineering, 2013, art. no. 162938, 2013

Simon, H. "A Behavioral Model of Rational Choice", in Models of Man, Social and Rational: Mathematical Essays on Rational Human Behavior in a Social Setting. New York: Wiley, 1957

Simon, H. "A mechanism for social selection and successful altruism". Science, Vol.250,No.4988,pp.1665–8, 1990

Simon, H. "Bounded Rationality and Organizational Learning", Organization Science,Vol.2, No.1,pp.125–134, 1991

Skinner, B. F. "Superstition' in the Pigeon", Journal of Experimental Psychology, Vol.38,No.2,pp.168–172, 1948

Spohn, W. "The Many Facets of the Theory of Rationality", Croatian Journal of Philosophy, Vol.2,pp.247–262, 2002

Stanciu-Viziteu, L.D. "The shark game: Equilibrium with bounded rationality", Lecture Notes in Economics and Mathematical Systems, Vol.662, pp. 103-111, 2012

Stapleton, T.J. "'They No Longer Care for Their Chiefs': Another Look at the Xhosa Cattle-Killing of 1856-1857', The International Journal of African Historical Studies, 24(2), pp. 383-392, 1991.Steuer, R.E. "Multiple Criteria Optimization: Theory, Computation and Application", New York: John Wiley, 1986

Szurley, J., Bertrand, A., Moonen, M. "On the use of time-domain widely linear filtering for binaural speech enhancement", IEEE Signal Processing Letters art. no. 6511977, Vol.20,No.7,pp. 649-652, 2013

Tettey T, Nelwamondo F. V. and Marwala T. "HIV data analysis via rule extraction using rough sets", In Proceedings of the 11th IEEE International Conference on Intelligent Engineering Systems, Budapest, Hungary, pp. 105-110, 29 June-1July 2007

Tim T, "Predicting HIV status using neural networks and demographic factors", University of the Witwatersrand MSc Dessertation,2007

Tim T.N. and Marwala T. "Computational Intelligence Methods for Risk Assessment of HIV", In Imaging the Future Medicine, Proceedings of the IFMBE, Vol. 14, Eds. Sun I. Kim and Tae Suk Sah, Springer-Verlag, Berlin Heidelberg, 2006, pp. 3581-3585

Tisdell, C. "Bounded Rationality and Economic Evolution: A Contribution to Decision Making, Economics, and Management", Cheltenham, UK: Brookfield, 1996

Tsang, E.P.K. "Computational intelligence determines effective rationality". International Journal on Automation and Control, Vol.5,No.1,pp. 63–6, 2008

Vapnik V. "The Nature of Statistical Learning Theory", Springer, 1995

Vilakazi B.C. and Marwala T. "Application of feature selection and fuzzy ARTMAP to intrusion detection", In Proceedings of the IEEE International Conference on Systems, Man and Cybernetics, Taiwan, pp. 4880-4885, 2006

Vyse, S.A. "Believing in Magic: The Psychology of Superstition", Oxford, England: Oxford University Press, 2000

Weber, M. "Ueber einige Kategorien der verstehenden Soziologie.", in Gesammelte Aufsaetze zur Wissenschaftslehre, Pp. 427-74 ,1922

Welsh, F. "A History of South Africa", HarperCollins, 2000.

Wigner E. P. "On the quantum correlation for thermodynamic equilibrium", Phys. Rev., vol. 40, pp. 749–759, 1932

Yang, W., Tavner, P.J., Court, R. "An online technique for condition monitoring the induction generators used in wind and marine turbines", Mechanical Systems and Signal Processing, Vol.38,No.1, pp. 103-112, 2013

Yao, J., Li, D. "Bounded rationality as a source of loss aversion and optimism: A study of psychological adaptation under incomplete information", Journal of Economic Dynamics and Control, Vol.37,No.1, pp. 18-31, 2013

Zadeh, L. A. "Fuzzy Sets, Fuzzy Logic, Fuzzy Systems", World Scientific Press, 1996

Zhang, X., Raza, A.S., Hood, D.C. "Detecting glaucoma with visual fields derived from frequency-domain optical coherence tomography", Investigative Ophthalmology and Visual Science, Vol.54, No.5, pp. 3289-3296, 2013

Zhang, Z., Aoki, Y., Toda, H., Imamura, T., Miyake, T. "Real world source separation by combining ICA and VD-CDWT in time-frequency domain", International Journal of Innovative Computing, Information and Control, Vol.9,No.4, pp. 1737-1757, 2013

Zimroz, R., Bartkowiak, A. "Two simple multivariate procedures for monitoring planetary gearboxes in non-stationary operating conditions", Mechanical Systems and Signal Processing, Vol.38, No.1, pp. 237-247, 2013

Zionts, S.; Wallenius, J. "An Interactive Programming Method for Solving the Multiple Criteria Problem", Management Science, Vol.22, No.6, pp.652–663,1976